\documentclass[conference]{IEEEtran}
\IEEEoverridecommandlockouts
\usepackage{cite}
\usepackage{amsmath,amssymb,amsfonts}
\usepackage{graphicx}
\usepackage{textcomp}
\usepackage{xcolor}
\usepackage{subcaption}
\usepackage{longtable}
\usepackage{booktabs}
\usepackage{microtype}
\usepackage{balance}
\usepackage[ruled,vlined]{algorithm2e}
\usepackage{mathrsfs,enumitem}
\usepackage{mathtools}
\usepackage{adjustbox}
\usepackage[draft]{todonotes}
\usepackage{multirow}
\usepackage{hyperref}
\def\BibTeX{{\rm B\kern-.05em{\sc i\kern-.025em b}\kern-.08em
    T\kern-.1667em\lower.7ex\hbox{E}\kern-.125emX}}

\SetCommentSty{mycommfont}

\title{Feedback Learning for Improving the Robustness of Neural Networks
}

\author{\IEEEauthorblockN{Chang Song}
\IEEEauthorblockA{\textit{Department of ECE} \\
\textit{Duke University}\\
Durham, NC, USA  \\
chang.song@duke.edu }
\and
\IEEEauthorblockN{Zuoguan Wang}
\IEEEauthorblockA{\textit{Black Sesame Technologies Inc}\\
Santa Clara, CA USA  \\
zuoguan.wang@bst.ai }
\and
\IEEEauthorblockN{Hai Li}
\IEEEauthorblockA{\textit{Department of ECE} \\
\textit{Duke University}\\
Durham, NC, USA  \\
hai.li@duke.edu }
}

\begin{document}

\maketitle

\begin{abstract}
Recent research studies revealed that neural networks are vulnerable to adversarial attacks. 
State-of-the-art defensive techniques add various adversarial examples in training to improve models' adversarial robustness. 
However, these methods are not universal and can't defend unknown or non-adversarial evasion attacks. 
In this paper, we analyze the model robustness in the decision space. 
A \textit{feedback learning} method is then proposed, to understand how well a model learns and to facilitate the retraining process of remedying the defects.
The evaluations 
according to a set of distance-based criteria show that our method can significantly improve models' accuracy and robustness against different types of evasion attacks. 
Moreover, we observe the existence of \textit{inter-class inequality} and propose to compensate for it by changing the proportions of examples generated in different classes.
\end{abstract}


\section{Introduction}
\label{intro}

In the past decade, the broad applications of deep learning techniques are the most inspiring advancements of machine learning~\cite{lecun2015deep}.
Compared to early attempts on neural networks~\cite{williams1986learning, lecun1998gradient}, modern deep learning models introduce more layers with complex structures and nonlinear transformations to model a high-level abstraction of data~\cite{bengio2013representation}.
The ability to learn from examples makes deep learning particularly attractive to cognitive applications, such as image and speech recognition~\cite{krizhevsky2012imagenet, hinton2012deep}, object detection~\cite{girshick2014rich}, natural language processing~\cite{collobert2008unified}, etc.
Moreover, deep learning systems demonstrated fascinating performance in many real-world applications and achieved near or even beyond human-level accuracy in solving classification problems in such domains, including handwritten digit recognition~\cite{lecun1998mnist}, image classification~\cite{graham2014fractional}, semantic scene understanding~\cite{wu2016apesnet}, etc.

The security industry has also adopted machine learning techniques in its practices~\cite{lian2007image, zhou2016image, zhang2015cross}.
Many of these applications are based on the strong classification capability of the learning models, including surveillance~\cite{tao2006human}, authentication~\cite{lian2007image}, facial recognition~\cite{taigman2014deepface}, vehicle detection~\cite{zhou2016image}, and crowd behavior analysis~\cite{zhang2015cross}.

However, recent research discovered that machine learning and neural network models are susceptible to \textit{adversarial attacks}, which apply small perturbations on input samples to fool models \cite{szegedy2013intriguing}.
Such attacks generally downgrade models' confidence levels on inputs and even result in misclassifications~\cite{goodfellow2014explaining}.
The amplitude of the perturbation that is used in adversarial attacks (a.k.a. \textit{adversarial strength}) can be quite small or even imperceptible to the human eyes. 
Furthermore, Papernot \emph{et al.} discovered that an elaborately-perturbed example (i.e., adversarial example) is transferable: the adversarial example crafted by a substitute model can not only deceive itself but also influence other models (e.g., victim models), even without knowing the internal structures and parameters of these victim models~\cite{papernot2017arxiva}.
These properties raise severe concerns on the security of deep learning technique.

A lot of research efforts have been put on adversarial examples and adversarial attacks~\cite{yuan2017adversarial}. 
Most works focused on maximizing the classification error (for attacks)/accuracy (for defenses) while minimizing the difference/distance between adversarial examples and original samples~\cite{carlini2017towards, szegedy2013intriguing, papernot2017practical, papernot2017arxiva}. 
As research are going further and deeper, the robustness of neural networks emerges as the new focus. 
Adversarial attacks are categorized as one type of \textit{evasion attacks}, which fool neural networks by introducing deliberately-modified examples at test time. 
The ultimate goal of improving model robustness is defending not only adversarial attacks, but any types of evasion attacks that attackers may conduct. 
One possible solution and explanation are based on the decision space analysis. 
We summarize the adversarial-based and decision-space-based related works in Section \ref{related-works}. 

Compared to the existing works on neural network security problems, especially on the adversarial attack and defense schemes, our major contributions in this work can be summarized as follows:

\vspace{-0.5mm}
\begin{itemize}[leftmargin=*]
	\item We prove the relation between the model robustness and margins (distances between samples and decision boundaries) in the decision space;
    \item We analyze the model robustness in the decision space based on an existing work \cite{he2018decision} and propose a set of distance-based criteria to evaluate it;
    \item We find out inter-class inequality exists in all datasets discussed in this work and class robustness can be utilized to improve model's overall robustness; 
    \item We propose a universal feedback learning method with inter-class inequality compensation to facilitate model retraining; 
    \item We show our method is effective in improving model accuracy and robustness against multiple types of evasion attacks with experiments on MNIST and CIFAR-10 datasets.
    \vspace{-4mm}
\end{itemize}

The remainder of this paper is organized as follows: 
Section~\ref{related-works} summarizes the existing works about adversarial attacks and decision-space-based methods;
Section~\ref{motivation} introduces the motivation of this work; 
Section~\ref{method} presents the details of our proposed method together with theoretical proof on method effectiveness; 
Section~\ref{case-study} discusses the experimental results; 
At last, we conclude this work in Section~\ref{conclusions}.
\section{Related Works}
\label{related-works}

\subsection{Adversarial Related Topics}

Kurakin \emph{et al.} \cite{Kurakin2017ICLR} showed that combining small batches of both adversarial examples and original data in adversarial training could make the model more resilient to adversarial attacks.
Carlini and Wagner \cite{carlini2017towards} demonstrated that defensive distillation does not significantly enhance the robustness of neural networks in some scenarios by introducing three new attack algorithms.
Cisse \emph{et al.} \cite{cisse2017parseval} introduced a layer-wise regularization method to reduce the neural network's sensitivity to small perturbations, which are difficult to be visually caught.

These adversarial-based approaches usually are effective for one or several specific adversarial attacks/defenses, but still vulnerable to different types or new adversarial attacks/defenses methods. 
For example, Athalye \emph{et al.} claimed that the attacks based-on their newly-proposed obfuscated gradients (which is a kind of gradient masking method) could circumvent 7 out of 9 noncertified white-box-secure defenses, all of which were accepted by \textit{International Conference on Learning Representations (ICLR) 2018} \cite{athalye2018obfuscated}. 
This shocking fact alerted the researchers in the area about the importance of universality for both attack and defense methods.

\subsection{Decision Related Topics}
\label{decision_related}

Adopted from the statistics community, a decision space (a.k.a an input space) refers to a vector space where all input samples lie. 
Decision boundaries are hyper-surfaces that partition the decision space into separate sets. 
During the learning of decision boundaries, neural networks attempt to minimize the empirical error, while support vector machines (SVMs) tend to maximize the empirical margin between the decision boundary and input samples~\cite{boser1992training}.

Besides the common MinMax-based adversarial researches, the latest studies shift the focus to the meaning of adversarial-related problems in decision spaces directly.
Tram{\`e}r \textit{et al.} \cite{tramer2017space} introduced methods of finding multiple orthogonal adversarial directions and showed that these perturbations span a multidimensional contiguous space of misclassified points. 
They believe that the higher the dimensionality of adversarial perturbations is, the more likely the subspaces of two models will intersect. 
Brendel \emph{et al.} introduced the \textit{boundary attack}, a decision-based attack that starts from a large adversarial perturbation and then seeks to reduce the perturbation while staying adversarial~\cite{brendel2018decision}.

Cao and Gong aimed to increase the robustness of neural networks~\cite{cao2017mitigating}. 
Instead of using only a test sample to determine which class it belongs to, hundreds of neighboring samples are generated in the surrounding hypercube and a voting algorithm is applied to decide the true label of the test sample. 
The results show that this straight-forward defense method is effective in improving neural networks' robustness in the sense of adversarial aspects without sacrificing the classification accuracy on legitimate examples. 
The method could substantially reduce the success rates of CW attacks~\cite{carlini2017towards} while other defense methods (such as adversarial training and distillation) couldn't.

Inspired by this region-based classification method \cite{cao2017mitigating}, He \emph{et al.} \cite{he2018decision} moved from hypercubes to larger neighborhoods. 
They proposed an orthogonal-direction ensemble attack called \textit{OptMargin}, which could evade the region-based classification defense mentioned in~\cite{cao2017mitigating}. 
They also analyzed margins (between samples and decision boundaries) and adjacent class information, then built a simple neural network to detect adversarial examples with these analysis results. 

Another state-of-the-art decision-based defense is proposed in~\cite{madry2017towards}. 
Madry \textit{et al.} found projected gradient descent (PGD) is a universal adversary among first-order approaches. To guarantee the adversarial robustness of the model, they re-started PGD from many points in the $l_{\infty}$ balls around data points to generate adversarial examples for training. 
The idea is intriguing, but it cannot provide a concrete guarantee against non-adversarial evasion attacks. 
In \cite{elsayed2018large}, a novel loss function is introduced to impose a margin on any layers of a model. 
To the best of the authors' knowledge, none of the existing works discovers or deals with inter-class inequality mentioned in the following Section \ref{motivation}.

\section{Motivation}
\label{motivation}

\begin{figure}[b]
	\centering
	\vspace{-4mm}
	\includegraphics[width=.85\linewidth]{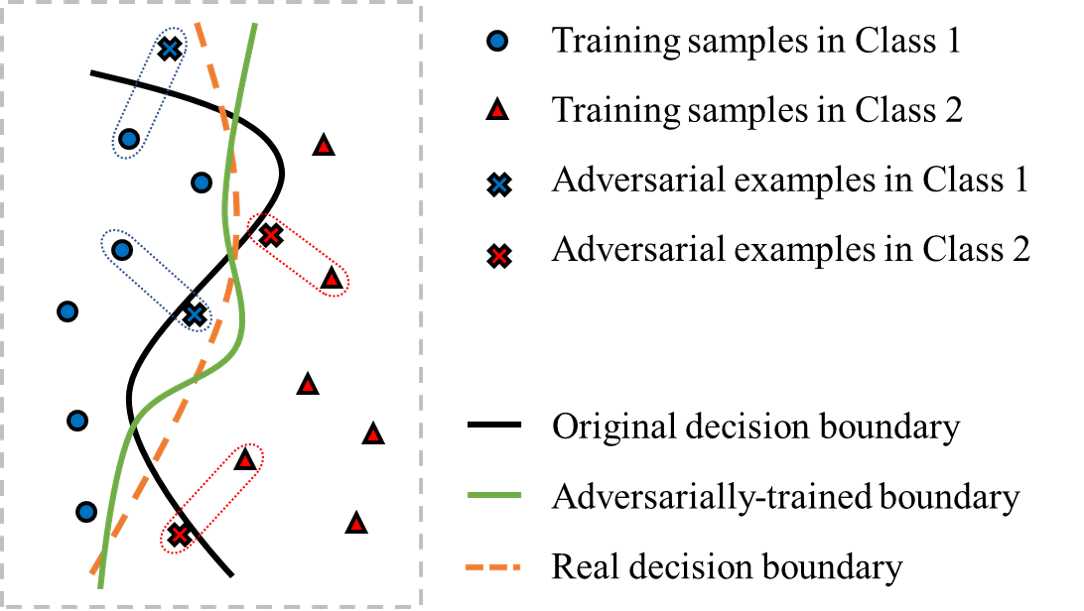}
	\caption{How do adversarial examples work in adversarial training.}
	\label{fig:boundary_push}
	\vspace*{-2mm}
\end{figure}

Inspired by Tram{\`e}r \emph{et al.}'s work \cite{tramer2017space}, we form an assumption on how generated examples facilitate the robustness improvement: 
generated examples can ``push" decision boundaries towards other classes, which enlarges the margins between real samples and decision boundaries. 
By retraining with generated examples in different directions, the margins of decision boundaries will enlarge. 
An illustration of this assumption with adversarial examples is given in Fig.~\ref{fig:boundary_push}.

Essentially, 
neural networks minimize the empirical error, while SVMs maximize the empirical margin between the decision boundary and training samples. 
These approaches have different objective functions, but target at the similar or even the same ultimate goal, that is, 
to classify the samples into correct classes as accurate as possible. 
Since most attacks are distance-based, the objective of model training shall include not only high accuracy but also large margins.



In many adversarial-related researches~\cite{carlini2017towards, szegedy2013intriguing, papernot2017practical, papernot2017arxiva}, the adversarial strengths are limited to certain ranges to guarantee the distortions are imperceptible. This limitation is equivalent to restricting the $l_{2}$ distance between original samples and adversarial examples. As defined, adversarial examples are supposed to cross decision boundaries and be misclassified with the minimum moving effort. If the model's decision boundaries can be trained to be far from every sample, then the generation of boundary-crossing examples (including adversarial examples) will be more difficult. As a result, the model's robustness against evasion attacks will be improved. This assumption will be further proved in Section~\ref{method}.

What's more, this kind of defense method is more generalized: not only is it beneficial to adversarial defenses, but also it can alleviate the damage of any distance-related evasion attack. 
In real-world scenarios, images captured by self-driving cars, security cameras, and webcams suffer from large noise and could totally vitiate classifiers. 
If the margin between samples and decision boundaries can be greatly increased, the model's robustness will be improved positively-correlated.

When observing all the margins and \textbf{adjacent/destination classes} (the first predicted class other than the original class as a sample keeps moving in any fixed direction), we discover another remarkable fact: adjacent classes are not equally-distributed, even though every class has the equal amount of training samples 
(6,000 training samples per class for MNIST~\cite{lecun1998gradient} and 5,000 for CIFAR-10~\cite{krizhevsky2009learning}).
In Table~\ref{tab:adjacent_proportions}, the percentages mark the proportions of total random search that fall into the corresponding adjacent class. 
As can be seen that class \textit{3} and \textit{Bird} are the most \textbf{robust (adjacent) classes} in MNIST and CIFAR-10, respectively, and class \textit{0} and \textit{Horse} are the most \textbf{vulnerable (adjacent) classes}. We name this phenomenon as \textbf{inter-class inequality}.

\begin{table*}[bt]
\centering
\caption{The proportions of adjacent classes show the inter-class inequality phenomenon. ``N/A" means the search ends with no adjacent class found. The higher the percentage, the more robust the adjacent class is.}
\label{tab:adjacent_proportions}
\resizebox{.9\textwidth}{!}{%
\begin{tabular}{lrrrrrrrrrrr} \toprule
\textbf{MNIST} & \textbf{0} & \textbf{1} & \textbf{2} & \textbf{3} & \textbf{4} & \textbf{5} & \textbf{6} & \textbf{7} & \textbf{8} & \textbf{9} & \textbf{N/A} \\
Adjacent & 0.02\% & 0.60\% & 1.59\% & 5.31\% & 4.52\% & 2.30\% & 1.07\% & 0.54\% & 4.29\% & 0.82\% & 78.95\% \\ \midrule
\textbf{CIFAR-10} & \textbf{Airplane} & \textbf{Automobile} & \textbf{Bird} & \textbf{Cat} & \textbf{Deer} & \textbf{Dog} & \textbf{Frog} & \textbf{Horse} & \textbf{Ship} & \textbf{Truck} & \textbf{N/A} \\
Adjacent & 0.20\% & 0.15\% & 44.61\% & 11.40\% & 7.51\% & 0.27\% & 25.62\% & 0.02\% & 2.47\% & 0.37\% & 7.39\% \\ \bottomrule
\end{tabular}
}
\vspace{-4mm}
\end{table*}

The proportional differences between vulnerable classes and robust classes are so tremendous that raises the question whether each class has the same robustness level. 
One hypothesis is that different classes have different spatial occupancy, which 
affect the construction of decision boundaries.
A robust class has a relatively large volume in the high-dimensional decision space, while a vulnerable class has a smaller volume region. 
As far as we observe from the results, vulnerable classes are adjacent to more classes, while robust ones are adjacent to fewer classes. 
If we can take class robustness into account, the model robustness can also be improved via improving class robustness. This will be further discussed in Section~\ref{method}.
\section{Feedback Learning and Theoretical Proof}
\label{method}

\setlength{\textfloatsep}{2pt}
\begin{algorithm}[t]
\DontPrintSemicolon
\SetAlgoLined
\SetNoFillComment
\LinesNotNumbered
\small
\caption{Feedback Learning Algorithm.}
\label{FL_Alg}
\SetAlgoNoEnd
\textbf{1. Training:}\tcp*{\textit{n} training samples.}

$model.train(samples, labels)$ 

\textbf{2. Sample Selection:}\tcp*{\textit{s} random samples.}

$selected\_index \leftarrow shuffle(arange(n))[:s]$ 

\textbf{3. Direction and Margin Calculation:}

\For {$i$ in $selected\_index$}{
    \For{$j$ in $random\_dirs$}{
        \For {$step=0$ to $max\_step$}{
            $new\_sample \leftarrow samples[selected\_index[i]]+step \times j$;
            
            $new\_label \leftarrow model(new\_sample)$;
            
            \If{$new\_label \neq labels[selected\_index[i]]$}{
                $[margin(i,j),adjacent(i,j)] \leftarrow [step,$\\$new\_label]$ \tcp*{Record results.}
                $break$;}}}}
\textbf{4. Example Generation:}

$count \leftarrow 0;$

\For {$k$ in $selected\_index$}{
$examples[count++] \leftarrow generate(samples[k],labels[k], $\\$margin[k],adjacent[k],random\_dirs);$}

$[retrain\_examples,retrain\_labels] \leftarrow shuffle([samples, labels],$\\$[examples,labels[selected\_index]]);$

\textbf{5. Retraining and Testing:}\\
\tcc{Retrain on the trained model.}
$model.train(retrain\_examples, retrain\_labels)$ 

$model.test(test\_samples, test\_labels)$ \tcp*{Test.}
    
\end{algorithm}

To alleviate the robustness problem and inter-class inequality mentioned in Section \ref{motivation}, we refine the boundary search originally presented by He \emph{et al.} \cite{he2018decision} and propose \textbf{feedback learning (F.L.)}. 
Its principle is to understand how well the model learns and generate the corresponding examples to facilitate the retraining process. 
The procedures of the complete feedback learning method can be divided into five steps: training, sample selection, direction and margin calculation (the boundary search), example generation, and retraining and testing. Algorithm \ref{FL_Alg} gives a demonstrative realization of the feedback learning algorithm.
In order to avoid confusion, we specify that ``\textbf{samples}" are legitimate, authentic, genuine samples collected from the real-world, and ``\textbf{examples}" are generated or perturbed instances. 
The detail about example generation will be discussed in Section \ref{example_selection}.

\subsection{Robustness Measurement}

First, we introduce the model's \textbf{mean margin matrix} $M$, of which the $i$-th row, $j$-th column element $m_{ij}$ is
\begin{equation}
\label{eq:1}
\small
m_{ij} = \frac{\sum_{y_{l}=j} \sum_{y_{k}=i}\left \| \eta_{k,l}\vec{d}_{k,l}\right \|_{2}}{N_{i,j}}=\frac{\sum_{y_{l}=j} \sum_{y_{k}=i} \eta_{k,l}}{N_{i,j}},
\end{equation}
%
where, $\eta_{k,l}$ and $\vec{d}_{k,l}$ respectively denote the margin and unit vector between samples and decision boundaries. 
All distances, margins mentioned in this paper are based on Euclidean metric. 
The numerator in the right hand side of Equation~(\ref{eq:1}) means the summed margins from all samples in class $i$ to the decision boundary between class $i$ and class $j$. $N_{i,j}$ is the total number of margins added for the origin and adjacent class pair $(i,j)$.

As discussed in previous section, we define the \textbf{model's robustness} as the margins between samples and decision boundaries. 
The relation can be formed as:
\begin{equation}
\label{eq:2}
\small
R = \sum_{j=1}^{c}\sum_{i=1}^{c} m_{ij},
\end{equation}
where $c$ is the total number of classes and $R$ denotes the robustness level of the model. 
A larger $R$ means the bigger overall margin and better robustness for the model.

Moreover, we define \textbf{class robustness} $R_{i}$ by:
\begin{equation}
\label{eq:3}
\small
R_{i} = \frac{\sum_{j=1}^{c} m_{ji}N_{j,i}}{\sum_{j=1}^{c} m_{ij}N_{i,j}}.
\end{equation}
Here we use summed margin to better express both margin and total number of traverses for each class, as there may exist cases with large $m_{ij}$ and small $N_{i,j}$ or small $m_{ij}$ and large $N_{i,j}$. In such cases, mean margin can not reflect how many traverses happen in the given experiment settings, while the summed margin can. 

According to Equation~(\ref{eq:3}), the class robustness of any class $i$ could be described as a ratio between total margins of class $i$ as adjacent class and total margins of class $i$ as origin class. 
The numerator measures the defensiveness of class $i$, and the denominator represents its offensiveness. 
If $R_{i}$ is greater than 1.0, we believe class $i$ is robust and mild. 
Otherwise, class $i$ is vulnerable if $R_{i}$ is less than 1.0. 
\textit{The class robustness $R_{i}$ can be used to measure the model's inter-class inequality and indicate which classes need further training.}

\subsection{Example Generation Criteria}
\label{example_selection}

Considering inter-class inequality of different classes, vulnerable classes have smaller margins and are prone to being transferred by other classes, while robust classes have relatively larger margins. Here, we utilize a straight-forward approach to improve the class robustness of vulnerable classes: \textit{by increasing the proportion of examples in vulnerable classes}. 
To be specific, all classes are categorized into three robustness levels, with three different settings for generating examples. 
If $R_{i}$ is among the top 20\% of all classes, it has ``high-level robustness" and only 20 samples in class $i$ are chosen for generating retraining examples (``minimum selection"). 
If $R_{i}$ is among the bottom 50\% in all classes, class $i$ has ``low-level robustness" and 150 samples will be utilized (``maximum selection"). 
Otherwise, class $i$ has ``medium-level robustness" and 100 samples will be utilized (``medium selection"). 

For each chosen sample, we generate retraining examples in 40 different directions with the top-40 minimum margins. The generation strength is $1.5\times - 2.0\times$ the margins we measured, which could guarantee boundary crossing (a theoretical proof on boundary retraining is provided in Section \ref{theoretical_derivation}). In this way, we construct a retraining dataset, shuffled with generated examples and all original training samples. Please note that all parameters mentioned here are empirical.

\subsection{Theoretical Derivation}
\label{theoretical_derivation}

\begin{figure}[t]
	\centering
	\includegraphics[width=.85\linewidth]{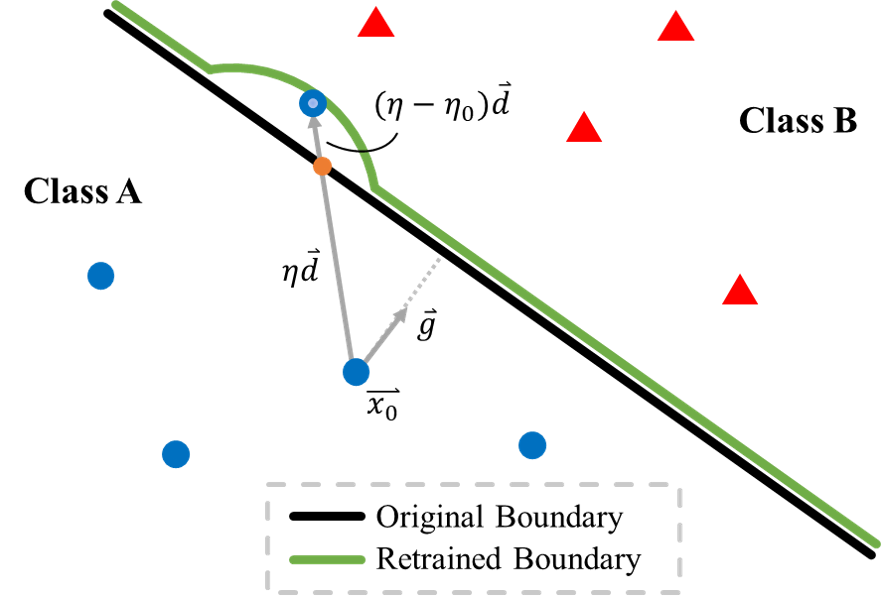}
	\caption{A binary classification scenario.}
	\label{fig:theory}
\end{figure}

To prove that retraining could positively influence decision boundaries, let us start with a binary classification problem illustrated in Figure \ref{fig:theory}.
Here, $\vec{x} \in \mathbb{R}^{D}$ is any $D$-dimensional input.
The binary classifier can be described as $f(\vec{x}, \vec{\theta})$, where $\vec{\theta}$ is the parameter. 
The decision boundary $f(\vec{x}, \vec{\theta})=0$ separates class A ($f(\vec{x}, \vec{\theta})<0$) and class B ($f(\vec{x}, \vec{\theta})>0$). 
Assume a training example $\vec{x}_{0}$ in class A with $f(\vec{x}_{0}, \vec{\theta})<0$. Unit vector $\vec{g}$ is the gradient direction from $\vec{x}_{0}$ towards $f(\vec{x}, \vec{\theta})=0$, and unit vector $\vec{d}$ is a random direction. 
The loss function $L(\vec{x})$ is:
\begin{equation}
\label{eq:4}
\small
L(\vec{x})=\frac{1}{2n} \sum _{\vec{x}} \left \| y(\vec{x})-f(\vec{x}, \vec{\theta}) \odot \sigma \right \|_{2} ^{2}. 
\end{equation}
Here, $n$ is the total number of samples trained, 
$y(\vec{x})
=\left\{\begin{matrix}
-1, \hspace{4pt} \text{if class A}\\ \hspace{10pt} 1, \hspace{4pt} \text{if class B}
\end{matrix}\right. 
$ is the label of samples, $\sigma$ is the activation function, and $\odot$ is the Hadamard product. 

\textit{Lemma 1}: If $\exists \eta_{0}>0, s.t. f(\vec{x}_{0}+\eta_{0}\vec{d}, \vec{\theta})=0$, then the direction $\vec{d}$ will have a positive contribution to the gradient $\vec{g}$, that is: $\cos \left \langle \vec{d}, \vec{g} \right \rangle>0$.

\textit{Proof}: As vector $\eta_{0}\vec{d}$ can be decomposed using vector resolution such as $\eta_{0}\vec{d}=\eta_{1}\vec{g}+\eta_{2}\vec{t}$, in which $\vec{t}$ is the tangent direction of $f(\vec{x}, \vec{\theta})=0$, $\vec{\eta_{1}}>0$. Thus:
\begin{equation}\label{eq:5}
\small
\begin{multlined}
\cos \left \langle \vec{d}, \vec{g} \right \rangle =\cos \left \langle \frac{\eta _{1}}{\eta _{0}}\vec{g}+\frac{\eta _{2}}{\eta _{0}}\vec{t}, \vec{g} \right \rangle= \\
\cos \left \langle \frac{\eta _{1}}{\eta _{0}}\vec{g}, \vec{g} \right \rangle+\cos \left \langle \frac{\eta _{2}}{\eta _{0}}\vec{t}, \vec{g} \right \rangle=\frac{\eta _{1}}{\eta _{0}}\left \|\vec{g} \right \|_{2}^{2}>0.
\end{multlined}
\end{equation}
After retraining with a generated example $\vec{x}_{e}=\vec{x}_{0}+\eta\vec{d}$, with $\eta>\eta_{0}$, we get a new boundary $f(\vec{x}, \vec{\theta})=0$.




\textit{Theorem 1}: After training with $\vec{x}_{e}$, the new classifier $f(\vec{x}, \vec{\theta}_{e})$ satisfies $f(\vec{x}_{e}, \vec{\theta}_{e}) < f(\vec{x}_{e}, \vec{\theta})$.


\textit{Proof}: The loss introduced by $\vec{x}_{e}$ is:
\begin{equation}\label{eq:6}
\small
\begin{multlined}
L(\vec{x}_{e})=\frac{1}{2} \left \| y(\vec{x}_{e})-f(\vec{x}_{e}, \vec{\theta}) \odot \sigma \right \|_{2}^{2} 
=\frac{1}{2} [1+f(\vec{x}_{e}, \vec{\theta}) \odot \sigma]^{2}.~~~~~~~~~~~~~
\end{multlined}
\end{equation}
In backpropagation, weights and biases are updated as:
\begin{subequations}
\begin{equation}\label{eq:7a}
\small
\begin{multlined}
\vec{w_{e}}=\vec{w}-\alpha \frac{\partial L(\vec{x_{e}})}{\partial \vec{w}} 
=\vec{w}-\alpha [f(\vec{x}_{e}, \vec{\theta}) \odot (\sigma + \sigma \sigma')] \frac{\partial f(\vec{x_{e}}, \vec{\theta})}{\partial \vec{w}},
\end{multlined}
\end{equation}
\begin{equation}\label{eq:7b}
\small
\begin{multlined}
\vec{b_{e}}=\vec{b}-\alpha \frac{\partial L(\vec{x_{e}})}{\partial \vec{b}}=\vec{b}-\alpha [f(\vec{x}_{e}, \vec{\theta}) \odot (\sigma + \sigma \sigma')].
\end{multlined}
\end{equation}
\end{subequations}
$\alpha > 0$ is the learning rate of the classifier and $\sigma'$ is the derivative of $\sigma$. We denote $\delta_{\sigma}=f(\vec{x}_{e}, \vec{\theta}) \odot (\sigma + \sigma \sigma')$.

Since $\vec{x}_{e}$ will be misclassified by the original boundary, we have $f(\vec{x}_{e}, \vec{\theta}) > 0$, then $\delta_{\sigma} > 0$. 

Suppose the binary classifier only has one layer, $f(\vec{x}, \vec{\theta})=\vec{w}\vec{x}+\vec{b}$. Then:
\begin{equation}\label{eq:8}
\small
\begin{multlined}
f(\vec{x}_{e}, \vec{\theta}_{e})-f(\vec{x}_{e}, \vec{\theta})=(\vec{w}_{e}-\vec{w})\vec{x}_{e}+(\vec{b}_{e}-\vec{b}).
\end{multlined}
\end{equation}
Based on Equation (\ref{eq:7a}) and Equation (\ref{eq:7b}), we can replace $\vec{w_{e}}$ and $\vec{b_{e}}$ in Equation (\ref{eq:8}) with $\vec{w}$ and $\vec{b}$:
\begin{equation}\label{eq:9}
\small
f(\vec{x}_{e}, \vec{\theta}_{e})-f(\vec{x}_{e}, \vec{\theta})=-\alpha \delta_{\sigma} (\left \| \vec{x}_{e} \right \|_{2}^{2}+1)<0.
\end{equation}
$f(\vec{x}, \vec{\theta})=\vec{w}^{(l)} [f_{l-1}(\vec{x}, \vec{\theta}_{l-1}) \odot \sigma]+\vec{b}^{(l)}$ if $f(\vec{x}, \vec{\theta})$ is a non-linear function. 
Here $l$ is the total number of layers of $f(\vec{x}, \vec{\theta})$, $\vec{w}^{(l)}$ and $\vec{b}^{(l)}$ are parameters in the last layer, $f_{l-1}$ is the representation of the first $(l-1)$ layers, and $\vec{\theta}_{l-1}$ is all parameters in $f_{l-1}$. Following similar steps as Equation (\ref{eq:9}):
\begin{equation}\label{eq:10}
\small
\begin{multlined}
f(\vec{x}_{e}, \vec{\theta}_{e})-f(\vec{x}_{e}, \vec{\theta})= \\
(\vec{w}_{e}^{(l)}-\vec{w}^{(l)})f_{l-1}(\vec{x}_{e}, \vec{\theta}_{l-1})+(\vec{b}_{e}^{(l)}-\vec{b}^{(l)}) 
=\\-\alpha \delta_{\sigma} (\left \| f(\vec{x}_{e}, \vec{\theta}_{l-1}) \right \|_{2}^{2}+1)<0.
\end{multlined}
\end{equation}
Therefore, we can guarantee that $f(\vec{x}_{e}, \vec{\theta}_{e})<f(\vec{x}_{e}, \vec{\theta})$ after retraining with $\vec{x}_{e}$ in the binary classifier $f(\vec{x}, \vec{\theta})$.

Now we consider multiclass classification problems. Learning directly from Moosavi-Dezfooli \emph{et al.} \cite{moosavi2016deepfool}, we know that mapping multiple labels is (or can be approximated as) a one-vs-all classification scheme. Moreover, our feedback learning method only focuses on one sample and one direction at a time. Two classes at most could be involved with any sample and  direction pair. Thus, it is always a binary classification problem in our case.

\begin{table*}[bt]
\centering
\caption{Mean margins of class center images (CCI) found in different CIFAR-10 models.}
\vspace*{-1mm}
\label{tab:center_image}
\resizebox{\textwidth}{!}{%
\begin{tabular}{llrrrrrrrrrrrr} \toprule
CCI & Model & 0 & 1 & 2 & 3 & 4 & 5 & 6 & 7 & 8 & 9 & Avg. & Std. \\ \cmidrule(lr){1-1} \cmidrule(lr){2-2} \cmidrule(lr){3-12} \cmidrule(lr){13-14}
\multirow{3}{*}{Ori.} & Ori. & 239.567 & 49.505 & 252.513 & 65.123 & 53.959 & 26.399 & 155.302 & 31.591 & 50.915 & 56.208 & 98.1082 & 85.681 \\
& Adv. & 27.161	& 75.020 & 245.622 & 50.273 & 79.537 & 45.759 & 255 & 66.710 & 89.535 & 56.269 & 99.089 & 81.728 \\
 & F.L. & 25.143 & 219.195 & 235.856 & 111.182 & 207.838 & 84.16 & 255 & 97.96 & 194.448 & 36.047 & 146.6829 & 85.323 \\
  & Reduced F.L. & 28.888 & 142.801 & 251.285 & 95.455 & 230.259 & 76.429 & 255 & 111.232 & 198.483 & 33.115 & 142.2947 & 86.766 \\
 \cmidrule(lr){1-1} \cmidrule(lr){2-2} \cmidrule(lr){3-12} \cmidrule(lr){13-14}
\multirow{3}{*}{F.L.} & Ori. & 26.615 & 49.505 & 248.739 & 37.857 & 53.959 & 16.805 & 82.651 & 26.325 & 47.715 & 39.919 & 63.009 & 67.783 \\
& Adv. & 89.823 & 75.020 & 233.650 & 52.883 & 79.537 & 67.908 & 254.547 & 57.907 & 83.937 & 98.880 & 109.409 & 72.486
 \\
 & F.L. & 150.723 & 219.195 & 255 & 132.132 & 207.838 & 117.963 & 255 & 107.617 & 206.296 & 190.074 & 184.1838 & 54.176 \\ 
  & Reduced F.L. & 219.107 & 142.801 & 176.329 & 171.791 & 230.259 & 106.657 & 255 & 91.034 & 190.249 & 137.5 & 172.0727 & 53.460 \\ \bottomrule
\end{tabular}
}
\vspace{-5mm}
\end{table*}

\vspace{-1mm}
\section{Case Study}
\label{case-study}


To prove the effectiveness of our proposed method, experiments were conducted with two models on two datasets: a Convolutional Neural Network (CNN) on MNIST and a ResNet \cite{he2016deep} on CIFAR-10. For both settings, we randomly picked 1,500 samples from training sets and performed boundary searches in orthogonal directions. We adopted the linear search method from He \emph{et al.} \cite{he2018decision} to find margins and adjacent classes: For MNIST, 784 random orthogonal directions were searched with step size 0.02 (the dimension of a MNIST sample is 784 and pixel values are 0 to 1) in both positive and negative directions. For CIFAR-10, 1,000 random orthogonal directions were searched with step size 2.0 (the dimension of a CIFAR-10 sample is 3,072 and pixel values are 0 to 255). We trained and retrained our models with parameter settings derived from Madry \emph{et al.} \cite{madry2017towards} on Tensorflow (v1.8.0)~\cite{tensorflow2015-whitepaper}. Most of the attacks we performed were using the Cleverhans library (v2.1.0)~\cite{papernot2018cleverhans}.

\begin{figure}[tbp]
	\centering
    \includegraphics[width=0.9\linewidth]{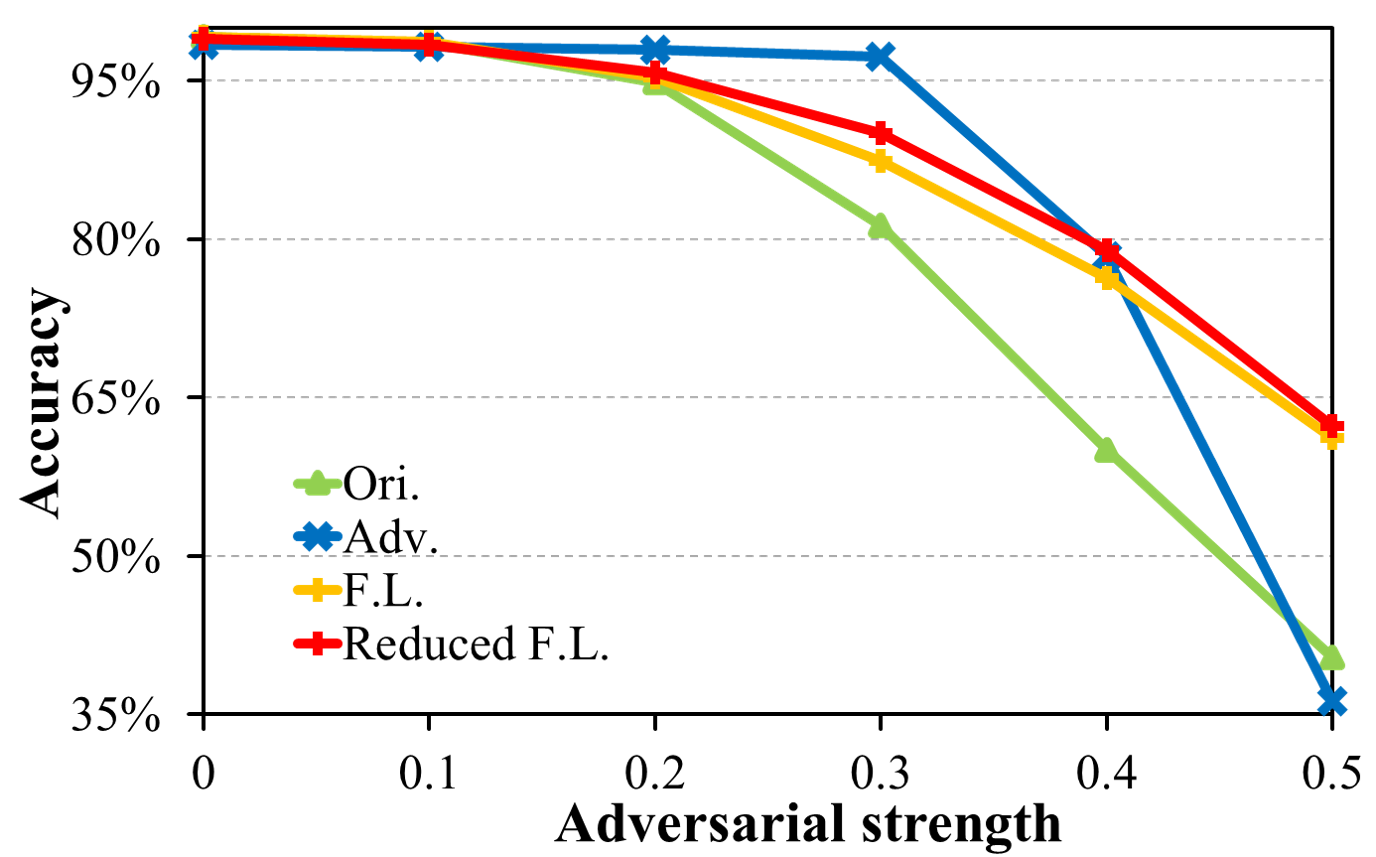}
	\caption{Accuracy curves of interested models under FGSM attacks on MNIST.}
	\label{fig:effectiveness}
\end{figure}

In Figure \ref{fig:effectiveness} and the rest of this paper, we will use the following  abbreviations for the models: \textbf{\textit{Ori.}} is the original one, \textbf{\textit{Adv.}} was adversarially-trained by Madry \emph{et al.} \cite{madry2017towards}, \textbf{\textit{F.L.}} was retrained from the original model using feedback learning with compensation to inter-class inequality, and \textbf{\textit{Reduced F.L.}} was retrained without considering inter-class inequality. 
The results show that feedback learning, both with or without inequality compensation, achieve better results than original training under FGSM attacks~\cite{goodfellow2014explaining}. 
For adversarial strength less than 0.4, the state-of-the-art adversarial model is the most accurate, while feedback learning surpasses it when the perturbation further increases. 
It is widely accepted that adversarial strength of 0.3 is large enough for any adversarial attacks on MNIST dataset. 
Benefiting from the randomness in direction selection, feedback learning can guarantee larger margins for more general directions and defend against severe attacks. 
The \textbf{\textit{Adv.}} model performs even worse than the \textbf{\textit{Ori.}} model at adversarial strength 0.5.
In fact, the defense of the \textbf{\textit{Adv.}} model degrades as the attack strength increases and when attack and defense methods differ. 
Our proposed feedback learning, in contrast, can defend adversarial-based attacks, even it does not use any adversarial-related information.

\vspace{-2mm}
\subsection{Margin Improvement and Inter-class Inequality Mitigation}

In Table~\ref{tab:center_image}, \textbf{class center images} (CCIs) are defined as examples with the largest mean margins in each class. First, we searched over the 1,500-sample set to find one CCI for each class and two different CCI sets were found based on two models (because the boundaries and margins are different in different models). 
Then we applied a linear search to measure the margins from CCIs to the decision boundaries of four models: \textbf{\textit{Ori.}}, \textbf{\textit{Adv.}}, \textbf{\textit{F.L.}}, and \textbf{\textit{Reduced F.L.}}.
The results are concluded as follows:

\begin{itemize}[leftmargin=*]
	\item The \textbf{\textit{Ori.}} model has the worst mean margin performance for both CCI sets. 
	The \textbf{\textit{F.L.}} model has the largest mean margins. 
	The overall mean margins of the \textbf{\textit{Adv.}} model is marginally larger than that of the \textbf{\textit{Ori.}} model. 
	The \textbf{\textit{F.L.}} model has highest overall mean margins and low standard deviations in both CCI cases, followed by the \textbf{\textit{Reduced F.L.}} model with slightly worse performances.
	\item In comparing CCI sets, the \textbf{\textit{F.L.}}'s CCI set has the largest mean margins (except on the \textbf{\textit{Ori.}} model) and the smallest standard deviations on all four models.
\end{itemize}

The results demonstrate that the retraining with cross-boundary examples always improve margin robustness, which is in consist with the proof in Section \ref{theoretical_derivation}. 
Employing feedback learning method additionally alleviates the inter-class inequality problem by increasing the margins in vulnerable directions.

\begin{table}[b]
\vspace*{2mm}
\centering
\caption{Model robustness against different attacks.} 
\vspace*{-1mm}
\label{tab:accuracy}
\resizebox{\linewidth}{!}{%
\begin{tabular}{lrrrrrr}\toprule
\multirow{2}{*}[-0.2em]{Attacks} & \multicolumn{3}{c}{MNIST} & \multicolumn{3}{c}{CIFAR-10} \\ \cmidrule(lr){2-4} \cmidrule(lr){5-7}
& Ori. & Adv. & F.L. & Ori. & Adv. & F.L. \\ \midrule
FGSM & 50.50\% & 97.57\% & 76.96\% & 36.60\% & 83.70\% & 56.20\% \\
CW-L2 & 39.40\% & 94.50\% & 51.63\% & 9.30\% & 54.20\% & 20.30\% \\
PGD & 18.62\% & 98.03\% & 64.54\% & 38.90\% & 85.20\% & 60.80\% \\
BIM & 15.36\% & 97.95\% & 56.39\% & 31.20\% & 85.10\% & 51.20\% \\
MIM & 38.08\% & 97.94\% & 76.39\% & 44.20\% & 85.10\% & 60.30\% \\
DeepFool & 32.92\% & 64.36\% & 44.75\% & 36.60\% & 83.70\% & 56.20\% \\ \midrule
Random & 78.19\% & 31.09\% & \textbf{91.07\%} & 45.40\% & 84.20\% & \textbf{87.50\%} \\
Gaussian & 77.31\% & 34.77\% & \textbf{90.65\%} & 79.14\% & 86.59\% & \textbf{90.58\%} \\ \bottomrule
\end{tabular}%
}
\end{table}

\subsection{Robustness Improvement and Decision Space Analysis}


Table~\ref{tab:accuracy} compares the accuracy performance of \textbf{\textit{Ori.}}, \textbf{\textit{Adv.}}, and \textbf{\textit{F.L.}} models on multiple evasion attack methods: fast gradient sign method (FGSM)~\cite{goodfellow2014explaining}, Carlini \& Wagner (CW-L2)~\cite{carlini2017towards}, projected gradient descent (PGD)~\cite{madry2017towards}, basic iterative method (BIM)~\cite{Kurakin2017ICLR}, momentum iterative method (MIM)~\cite{dong2017boosting}, DeepFool~\cite{moosavi2016deepfool}, random noise (with a uniform distribution), and Gaussian noise. Please note that all attack parameters were carefully tuned for better effectiveness display (so that none of the models should perform worse than a random guess). For adversarial-related attacks, the \textbf{\textit{Adv.}} model is the most accurate overall, while feedback learning improves model robustness to a certain extent. As to random and Gaussian noise attacks, feedback learning clearly reaches the best accuracies on both datasets, while the \textbf{\textit{Adv.}} model's robustness could be destroyed against non-adversarial attacks (especially on low-resolution datasets such as MNIST).

The result shows our method can improve accuracies on both adversarial-based examples and non-adversarial examples, compared to the \textbf{\textit{Ori.}} model. The reason why the \textbf{\textit{F.L.}} model doesn't achieve better result than the \textbf{\textit{Adv.}} model on adversarial-based examples is that the latter is trained especially for these kinds of attacks. As we can see that the \textbf{\textit{Adv.}} model has worse results on non-adversarial examples than the \textbf{\textit{F.L.}} model.

\begin{figure*}[btp]
	\centering
	\begin{subfigure}{0.49\linewidth}
		\centering
		\includegraphics[width=\textwidth]{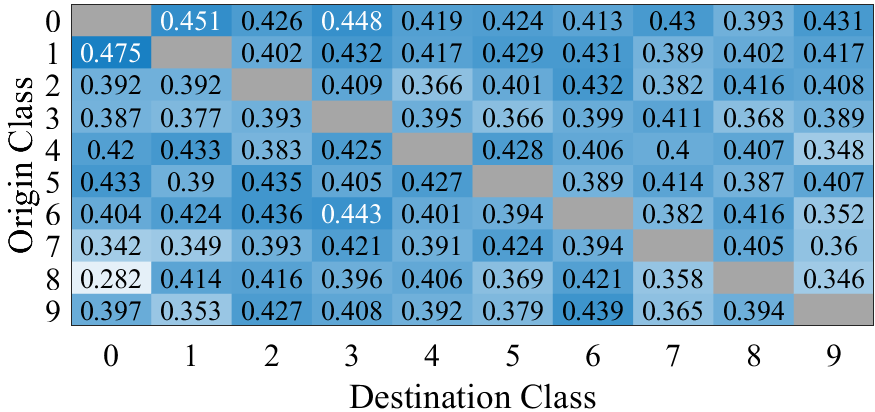}
		\caption{\textbf{\textit{Ori.}} MNIST model.}
		\label{fig:ori_heatmap_mnist}
	\end{subfigure}
	\begin{subfigure}{0.49\linewidth}
		\centering
		\includegraphics[width=\textwidth]{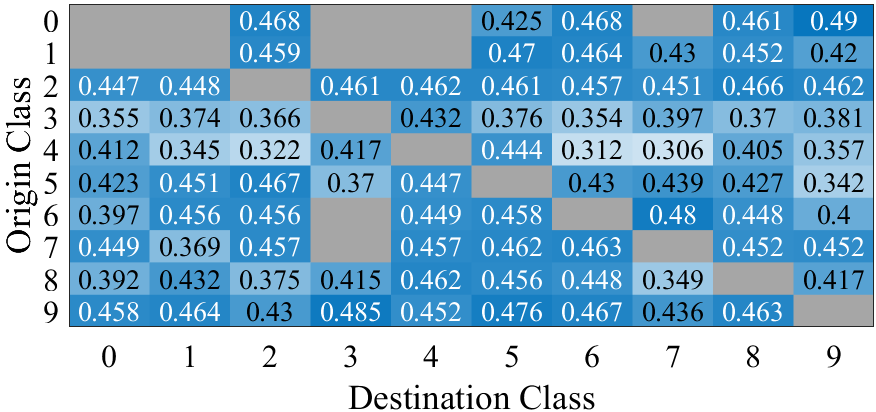}
		\caption{\textbf{\textit{F.L.}} MNIST model.}
		\label{fig:retrained_heatmap_mnist}
	\end{subfigure}
	\begin{subfigure}{0.49\linewidth}
		\centering
		\includegraphics[width=\textwidth]{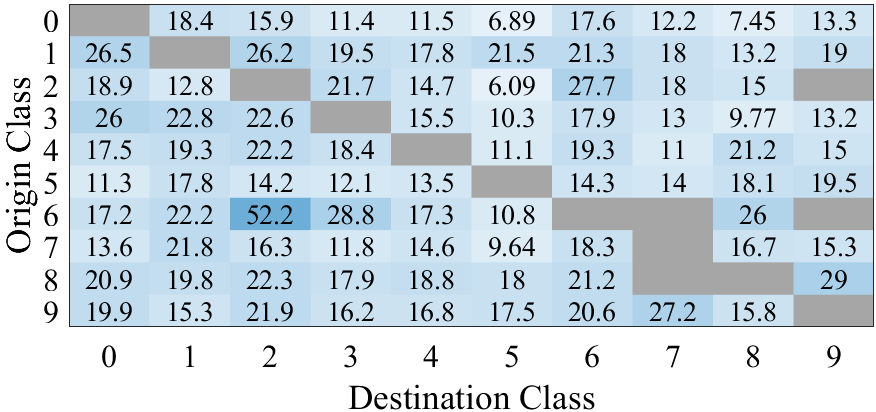}
		\caption{\textbf{\textit{Ori.}} CIFAR-10 model.}
		\label{fig:ori_heatmap_cifar}
	\end{subfigure}
	\begin{subfigure}{0.49\linewidth}
		\centering
		\includegraphics[width=\textwidth]{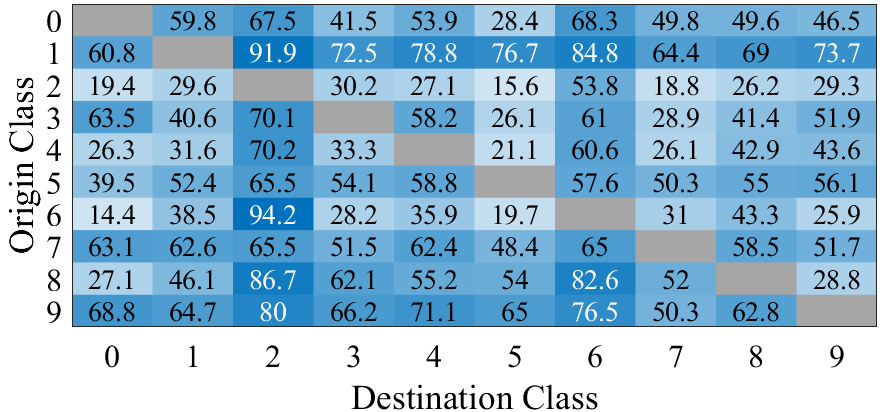}
		\caption{\textbf{\textit{F.L.}} CIFAR-10 model.}
		\label{fig:retrained_heatmap_cifar}
	\end{subfigure}
	\vspace*{-2mm}
	\caption{Mean margin matrices comparisons between (a) \& (b) MNIST and (c) \& (d) CIFAR-10 models. Grey blocks mean no adjacent class exist for given origin and destination (adjacent) classes.}
	\label{fig:heatmaps}
	\vspace*{-2mm}
\end{figure*}

Furthermore, we measured and calculated the margins and robustness of all the four models. Please note that we omit the class ``N/A" in Table~\ref{tab:adjacent_proportions}, because the margin is large enough to be ignored if no adjacent class found. 
The heat maps of mean margin matrices in Figure~\ref{fig:heatmaps} give visual comparisons between the \textit{\textbf{Ori.}} and \textit{\textbf{F.L.}} models' margin measurements. 
Mean margin matrices of the \textit{\textbf{Adv.}} and \textit{\textbf{Reduced F.L.}} models are not shown due to page limit. 
Here, MNIST samples are scaled to $[0, 1]$ and CIFAR-10 samples are kept in original format $[0, 255]$ in pixel value. 
For MNIST dataset, as class ``3" and ``4" fall into ``high-level robustness" category and class ``0" and ``1" have ``low-level robustness" in the \textit{\textbf{Ori.}} model, more samples were generated for retraining in class ``0" and ``1" than class ``3" and ``4" as defined in Section \ref{example_selection}. 
After retraining with feedback learning, the mean margins of class ``3" and ``4" slightly decrease, while other classes' mean margins increase. 
We also note that more margins are beyond searching range (which is half of the pixel value range, 0.5 for MNIST and 127.5 for CIFAR-10) after retraining. 
These changes result in an overall improved summed margin and mean margin. For CIFAR-10 dataset, the improvement is more significant.

Moreover, we calculate the robustness levels:  $R_{Ori.}=36.15$,
$R_{Adv.}=31.74$, 
$R_{F.L.}=39.12$, and 
$R_{Reduced~F.L.}=35.52$ on MNIST; and $R_{Ori.}=2035.13$,
$R_{Adv.}=3969.10$, 
$R_{F.L.}=4634.36$, and 
$R_{Reduced~F.L.}=3306.59$ on CIFAR-10. 
Here, margins with no adjacent class are counted as the maximum searching range. 
The results prove that our feedback learning method substantially increases the margins from samples to decision boundaries and also ensure models have larger overall margins comparing those obtained from retaining without considering inter-class inequality. 
More analyses on causes of such results are given in Section \ref{class_robustness}.

\subsection{Class Robustness Comparison}
\label{class_robustness}

\begin{table*}[bt]
\centering
\caption{Class robustness of different models. For the \textbf{\textit{F.L.}} model, $\bigtriangleup$: Maximum selection for ``low-level robustness" classes, $\square$: Medium selection for ``medium-level robustness" classes, $\bigtriangledown$: Minimum selection for ``high-level robustness" classes.}
\label{tab:class_robustness}
\resizebox{0.9\textwidth}{!}{%
\begin{tabular}{lrrrrrrrrrrrr} \toprule
\textbf{MNIST} & 0 $\bigtriangleup$ & 1 $\bigtriangleup$ & 2 $\bigtriangleup$ & 3 $\bigtriangledown$ & 4 $\bigtriangledown$ & 5 $\square$ & 6 $\bigtriangleup$ & 7 $\bigtriangleup$ & 8 $\square$ & 9 $\bigtriangleup$ & Avg. & Std.\\ \cmidrule(lr){1-1} \cmidrule(lr){2-11} \cmidrule(lr){12-13}
$R_{i, Ori.}$ & 0.005 & 0.149 & 0.955 & 7.237 & 6.662 & 2.135 & 0.661 & 0.187 & 3.075 & 0.176 & 2.124 & 2.73 \\
$R_{i, Adv.}$ & 0.702 & 0.007 & 4.425 & 1.511 & 0.221 & 1.296 & 0.922 & 0.182 & 3.603 & 0.293 & 1.316 & 1.517 \\
$R_{i, F.L.}$ & 29.316 & 31.268 & 15.408 & 0.009 & 0.007 & 1.856 & 10.387 & 2.318 & 5.858 & 25.966 & 12.24 & 12.48 \\
$R_{i, Reduced \hspace{2pt} F.L.}$ & 3.665 & 32.208 & 0.041 & 0.553 & 128.07 & 0.809 & 0.388 & 1.227 & 1.049 & 0.001 & 16.80 & 40.32 \\ \midrule
\textbf{CIFAR-10} & 0 $\bigtriangleup$ & 1 $\bigtriangleup$ & 2 $\bigtriangledown$ & 3 $\square$ & 4 $\square$ & 5 $\bigtriangleup$ & 6 $\bigtriangledown$ & 7 $\bigtriangleup$ & 8 $\square$ & 9 $\bigtriangleup$ & Avg. & Std. \\ \cmidrule(lr){1-1} \cmidrule(lr){2-11} \cmidrule(lr){12-13}
$R_{i, Ori.}$ & 0.032 & 0.011 & 14.005 & 0.888 & 0.565 & 0.021 & 1.110 & 0.002 & 0.151 & 0.029 & 1.681 & 4.349 \\
$R_{i, Adv.}$ & 0.066 & 0.0074 & 1.768 & 0.104 & 1.240 & 0.013 & 25.46 & 0.299 & 0.031 & 0.497 & 2.948 & 7.93 \\
$R_{i, F.L.}$ & 0.020 & 0.059 & 9.147 & 0.343 & 0.573 & 0.023 & 11.209 & 0.015 & 0.231 & 0.052 & 2.167 & 4.254\\
$R_{i, Reduced \hspace{2pt} F.L.}$ & 0.721 & 0.894 & 14.628 & 0.057 & 0.128 & 0.121 & 1.559 & 0.470 & 0.197 & 0.066 & 1.884 & 4.503 \\ \bottomrule
\end{tabular}
}
\vspace*{-4mm}
\end{table*}

Table \ref{tab:class_robustness} compares class robustness of four models: \textbf{\textit{Ori.}}, \textbf{\textit{Adv.}}, \textbf{\textit{F.L.}} and \textbf{\textit{Reduced F.L.}}. The \textbf{\textit{Adv.}} model controls neither model robustness nor class robustness, thus it has the most unstable performance among all. Feedback learning improves overall class margins with acceptable expense: Several ``high-level" and ``medium-level robustness" classes may become more vulnerable (such as ``3" and ``4" for both datasets), while ``low-level robustness" classes have remarkably better robustness. Without considering inter-class inequality, the \textbf{\textit{Reduced F.L.}} model still achieves better robustness than the \textbf{\textit{Ori.}} model, but relatively worse than the \textbf{\textit{F.L.}} model in the sense of overall margins and standard deviations.

Why does the \textbf{\textit{F.L.}} model achieve better results than the \textbf{\textit{Reduced F.L.}} model, even when the latter employs more examples in retraining? 
This is because the total capacity of the decision space is constant, and decision boundaries are only borders to partition the given space. Retraining with any technique will only relocate boundaries and redistrict class regions. 
As the inter-class inequality problem exists, training each class with the same emphasis won't increase classes' robustness simultaneously, but may even result in severer maldistribution of class domains.
This also explains why $R_{Reduced~F.L.}$ is even smaller than $R_{Ori.}$ on MNIST. 
Feedback learning attempts to relax such an inequality by allowing vulnerable classes to make more contributions to the models, thus the boundary relocation is more controllable.

\subsection{Computation Complexity Analysis}

The boundary search is the most computation-consuming part in our feedback learning method. 
As the dimension of inputs increase, the computation cost will go up. 
The following two explanations/solutions could greatly alleviate this issue.

\begin{itemize}[leftmargin=*]
    \item The boundary search happens before the retraining process, which means it is conducted only once for each model. What's more, the margin information can be reused to further retrain the model multiple times until the model reaches the desired robustness performance;
    \item The boundary search in each direction for each sample can be highly parallel. This won't reduce the computation cost but will greatly shorten the searching time. Here we assume the major concern is model robustness and computation time, not the computing power. Thus, our method can be deployed on datasets with high dimensional inputs.
\end{itemize}
\section{Conclusions}
\label{conclusions}

In this work, we first analyze the model robustness in the decision space. According to it, we propose a feedback learning method to understand how well a model learns and facilitate the model's retraining to remedy the defects. A set of distance-based criteria for model robustness evaluation show that our method can significantly improve models accuracy and robustness against different types of attacks. Moreover, we observe the existence of inter-class inequality, which can be compensated by changing the proportions of examples generated in each class.

\balance
\bibliographystyle{unsrt}
\bibliography{ICMLA19.bib}

\end{document}